\documentclass[times, review, 10pt]{elsarticle}
\usepackage{geometry}
\usepackage{amssymb}

\usepackage{subfigure}
\usepackage{longtable}
\usepackage{multirow}
\usepackage{mathrsfs}
\usepackage{amsfonts}
\usepackage{amsmath}
\usepackage{array}
\usepackage{color}
\usepackage[ruled]{algorithm2e}
\usepackage{float}
\biboptions{numbers,sort&compress}

\journal{Pattern Recognition}

\begin{document}

\begin{frontmatter}
\title{Research on Violent Text Detection System Based on BERT-fasttext Model}
\author[firstauthor]{Yongsheng Yang}
\author[firstauthor2]{Xiaoying Wang}

\address[firstauthor]{School of Information Engineering, Zhongnan University of Economics and Law, Wuhan, 430073, China}
\address[firstauthor2]{School of Information Engineering, Zhongnan University of Economics and Law, Wuhan, 430073, China}

\begin{abstract}

In the digital age of today, the internet has become an indispensable platform for people's lives, work, and information exchange. However, the problem of violent text proliferation in the network environment has arisen, which has brought about many negative effects. In view of this situation, it is particularly important to build an effective system for cutting off violent text. The study of violent text cutting off based on the BERT-fasttext model has significant meaning. BERT is a pre-trained language model with strong natural language understanding ability, which can deeply mine and analyze text semantic information; Fasttext itself is an efficient text classification tool with low complexity and good effect, which can quickly provide basic judgments for text processing. By combining the two and applying them to the system for cutting off violent text, on the one hand, it can accurately identify violent text, and on the other hand, it can efficiently and reasonably cut off the content, preventing harmful information from spreading freely on the network. Compared with the single BERT model and fasttext, the accuracy was improved by 0.7\% and 0.8\%, respectively. Through this model, it is helpful to purify the network environment, maintain the health of network information, and create a positive, civilized, and harmonious online communication space for netizens, driving the development of social networking, information dissemination, and other aspects in a more benign direction.

\end{abstract}


\end{frontmatter}


\section{Introduction}

Violent text is a text expression that expresses violent tendencies, inflammatory or offensive content in the form of language. Its core feature is to convey threats, intentions to harm, hatred or intimidation to individuals or groups through text, which may cause negative psychological, emotional or social effects. Its characteristics include threats, hatred, or descriptions of violent acts against individuals or groups, language that incites others to commit violent acts, and the use of insults, slander, or discriminatory words to intensify conflicts, which may be directed at individuals or social groups, such as "just by being able to tweet this insufferable bullshit proves trump a nazi you vagina".\cite{Zhu2024TowardsAC} When studying or analyzing violent texts, it is usually necessary to combine natural language processing (NLP) technology with ethical considerations to identify and manage such content through classification models or sentiment analysis technology.\cite{Mukhamedshin2024NMRSO} The main contributions of this paper are summarized as follows: 

In addition to the simple big data analysis hotspot method, this project chooses a more accurate and more practical natural language processing technology. The information processed by relying on a large amount of data is of certain reference value, but there are a lot of duplications, and it is impossible to detect other branches of cyber violence centered on this hotspot.\cite{Wan2024BranchTB} While conducting an overall analysis, we process and locate the language of individuals, and through specific analysis of people's emotions, we dig out violent information at the language level. Reasonable application of word segmentation and stop words to accurately analyze the content, mainly using the BERT model to collect data for automatic feature extraction. The entire test process is optimized using dictionaries and rules to supplement the deviation of analyzing the topic of cyber violence based on keywords alone. This technology of analyzing and classifying emotions is more suitable for supervising subjective and negative cyber violence activities, so as to achieve the effect of timely discovery and interception of cyber violence.\cite{Capozzi2024OnTL}

\begin{figure*}[!t]
	\begin{center}
		\includegraphics[width=0.96\linewidth]{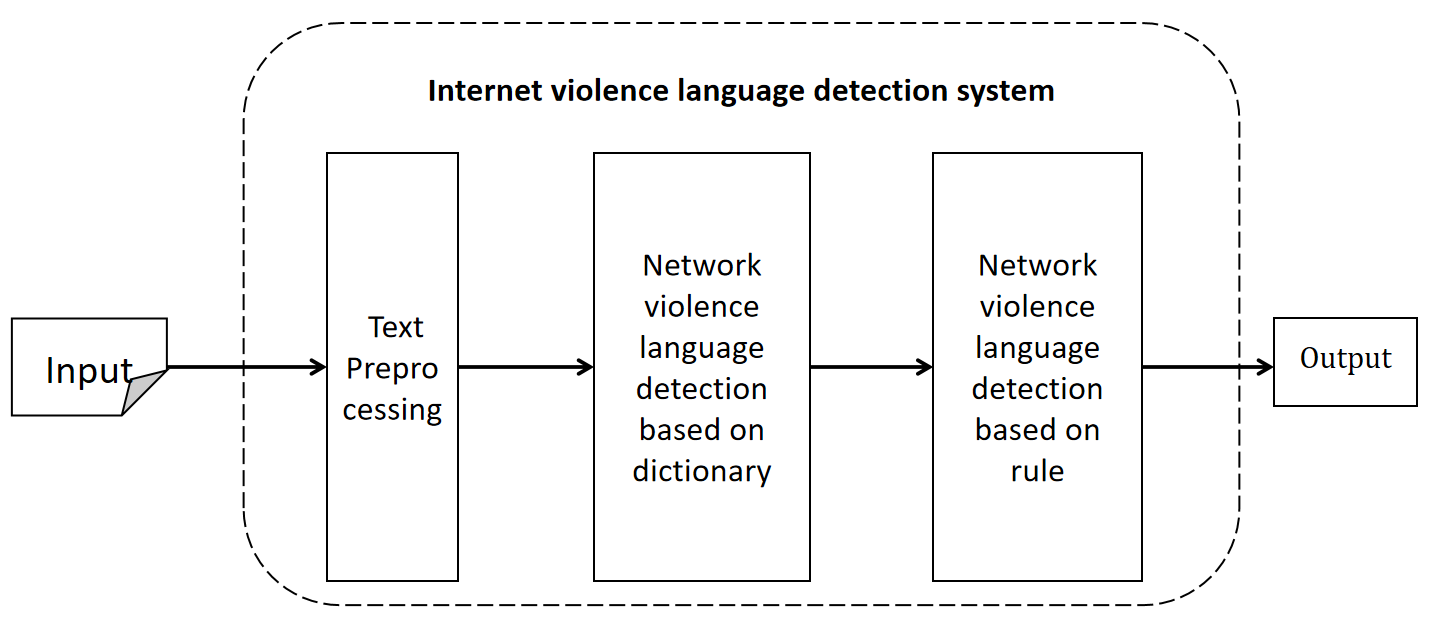}
	\end{center}
	\caption{Violent Text Detection System}
	\label{Violent Text Detection System}
\end{figure*}

The rest of this paper is organized as follows.In the second part, we will elaborate on the experience and methods of relevant industry professionals in this direction. In the third part, we will introduce the models involved and the measures of our method in detail, including the BERT model and the fasttext algorithm. In the fourth part, we will show our experimental evaluation indicators and compare them with other models to prove that our results are better.

\section{Related Work}
"Language violence" originally originated from the Western postmodern philosophy school \cite{Capozzi2024OnTL}. Foucault  believed that discourse influences, regulates and constrains our thoughts and behaviors, and discourse is composed of interrelated statements\cite{Yang2024CCOCRAC}. We generally believe that "language violence" is the use of discriminatory words such as slander, insults, abuse, and contempt to cause psychological and spiritual violations and harm to others\cite{Warner2024SmarterBF}. "Internet language violence" also mainly refers to language violence that occurs on the Internet, that is, a social phenomenon that uses language attacks to cause some degree of spiritual and psychological harm to the attacked object.\cite{Tao2024LLMsAA}

\subsection{\textbf{BERT}}

\begin{figure*}[!t]
	\begin{center}
		\includegraphics[width=0.96\linewidth]{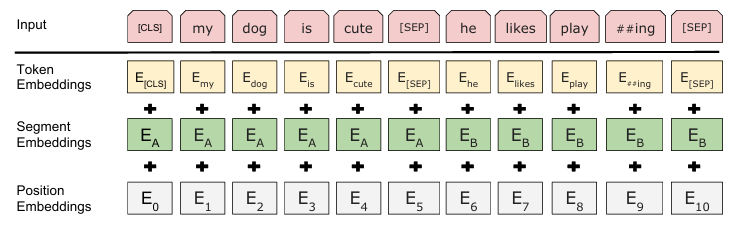}
	\end{center}
	\caption{ BERT input representation. The input embeddings are the sum of the token embeddings, the segmentation embeddings and the position embeddings.}
\end{figure*}

The BERT model replaces traditional RNN and CNN with the Attention mechanism, converting the distance between two words at any position to 1, effectively solving the problem of long-term dependency in NLP. \cite{Yang2024CCOCRAC}The input encoding vector of BERT (length 512) consists of the following three embedding features:
1. WordPiece embedding: Split words into limited common subword units to balance word effectiveness and character flexibility. For example, "playing" is split into "play" and "ing".
2. Position Embedding: Encode word position information into feature vectors to introduce the position relationship between words.
3. Segment Embedding: Used to distinguish sentence pairs. For example, in the question-answering scenario, the feature values of sentence A and sentence B are 0 and 1 respectively.
In addition, [CLS] indicates that the feature is used for classification tasks, and [SEP] is used to distinguish sentences.\cite{Sokli2024InvestigatingMO}
During training, BERT will randomly mask 15\% of WordPiece Tokens. The specific strategy is:

80\% is replaced with [Mask] (for example: my dog is hairy $\to$ my dog is [Mask]).

10\% are replaced with other words (e.g. my dog is hairy $\to$ my dog is apple).

10\% remain the same(e.g. my dog is hairy $\to$ my dog is hairy).

BERT performs multi-task learning through two self-supervised tasks:

1. Masked Language Model (MLM): predict the masked words (fill-in-the-blank task).\cite{Mukhamedshin2024NMRSO}
\begin{equation}
	P(w_i | w_{\neq i}) = \text{softmax}(W^t_0\cdot \text{BERT}(w_1, w_2, \dots, \text{[MASK]}, \dots, w_N))
\end{equation}

Where [MASK] represents the position of the masked word, BERT(w1, w2, ..., [MASK], ..., wN ) represents an
encoded representation of the input text sequence by the
BERT model.

2. Next Sentence Prediction (NSP): predict the contextual relationship between two sentences.\cite{Basem2024OptimizedQP}
\begin{equation}
	P(\text{IsNext} | A, B) = \sigma(W_c^t \cdot [\text{BERT}(A); \text{BERT}(B)])
\end{equation}

Where $\sigma$ is the sigmoid function,$\mathbf{W}_c^t$
is the weight matrix
of the classifier, and [BERT(A);BERT(B)] represents the
concatenation of the BERT encoded representations of the
sentences A and B.

\subsection{\textbf{Fasttext model}}
FastText, like Word2vec, was proposed by Mikolov. It is a technology that Facebook opened in 2016 for generating word vectors and performing text classification. Before there were word vector models including Word2vec, word vectors were represented using the bag-of-words model. However, generating word vectors using the bag-of-words model will result in dimensionality disaster and does not consider the order and semantic information of words. Now the mainstream Word2vec can represent words as vectors of a certain dimension without causing dimensionality disaster. The FastText model is similar to the CBOW model of Word2vec, but the CBOW model predicts the intermediate words based on the context, while the FastText model predicts the label based on the entire sequence.
\begin{figure*}[!t]
	\begin{center}
		\includegraphics[width=0.5\linewidth]{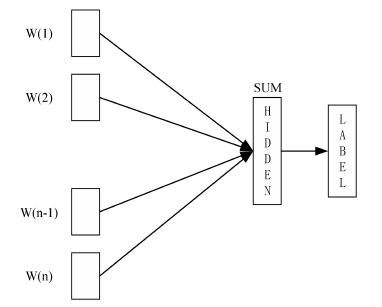}
	\end{center}
	\caption{ The structure of fasttext model. The input  will be added to hidden layer.And then they will be sent to label.}
\end{figure*}

Like Word2vec, the FastText model is also divided into three layers, namely the input layer, the hidden layer, and the output layer. The input layer is the word sequence $w_1, w_2, ..., w_n$ of the entire text, and then the word vectors of each word are accumulated and averaged, like $\frac{1}{n}  {\textstyle \sum_{1}^{n}}W_i $ , undergoing nonlinear transformation in the hidden layer, and finally outputting the label of the entire text. In the output layer, a Huffman tree is constructed by using labels and frequencies. Each leaf node in the Huffman tree represents a label, and each non-leaf node indicates that a binary classification is required here. The probability of the positive category is represented by $\sigma(X_i \theta)$, and the probability of the negative category can be represented by 1-$\sigma(X_i \theta)$. The specific formula for the positive category is as follows.
\begin{equation}
	\sigma(X_i \theta) = \frac{1}{1+e^{-X_i \theta} } 
\end{equation}

$X_i$ represents the feature vector. By performing sentiment classification on $X_i$, the probability of the predicted category label is set to $Y$, and the specific formula is as follows.
\begin{equation}
	P(Y_i | X_i) =  {\textstyle \prod_{j=2}^{l}} P(d_j | X_i, \theta_{j-1})
\end{equation}

Among them,
\begin{equation}
	P(d_j | X_i, \theta_{j-1})=P\left(\mathrm{~d}_{\mathrm{j}} \mid X_{i}, \theta_{j-1}\right)=\left\{\begin{array}{ll}
		\sigma\left(X_{\mathrm{i}} \theta\right), & \mathrm{d}_{j}=1 \\
		1-\sigma\left(X_{\mathrm{i}} \theta\right), & \mathrm{d}_{j}=1
	\end{array}\right.
\end{equation}

Then the log likelihood of formula above is calculated to obtain the formula.
\begin{equation}
	P(Y_i | X_i) =  {\textstyle \prod_{j=2}^{l}} P(d_j | X_i, \theta_{j-1})
\end{equation}

Finally, the objective function is obtained, as shown below.
\begin{equation}
	l = \frac{1}{n}{\textstyle \sum_{i=1}^{n}}\log_{}{P(Y|X)} 
\end{equation}

Then the parameters are adjusted by the gradient ascent method to maximize the value of the formula, which is consistent with the CBOW in the previous article. The difference between the FastText model and Word2vec is that the input layer of the FastText model is all the words in the entire text, while the input layer of Word2vec is only the upper and lower words of the target word in the window. The FastText model finally predicts the label of the entire text, that is, the probability that the text belongs to a certain category, while the CBOW model of Word2vec only predicts the target word. When Word2vec trains the text, it only adds the upper and lower word vectors, ignoring the order of words in the text. For example, the two sentences "Do you like to eat apples?" and "Do you not like to eat apples" will be treated as the same sentence in Word2vec. The FastText model solves this problem with n-grams. The FastText model adds n-gram vocabulary to the hidden layer.\cite{Fussell2024ComparingLL}

\begin{figure*}[!t]
	\begin{center}
		\includegraphics[width=0.96\linewidth]{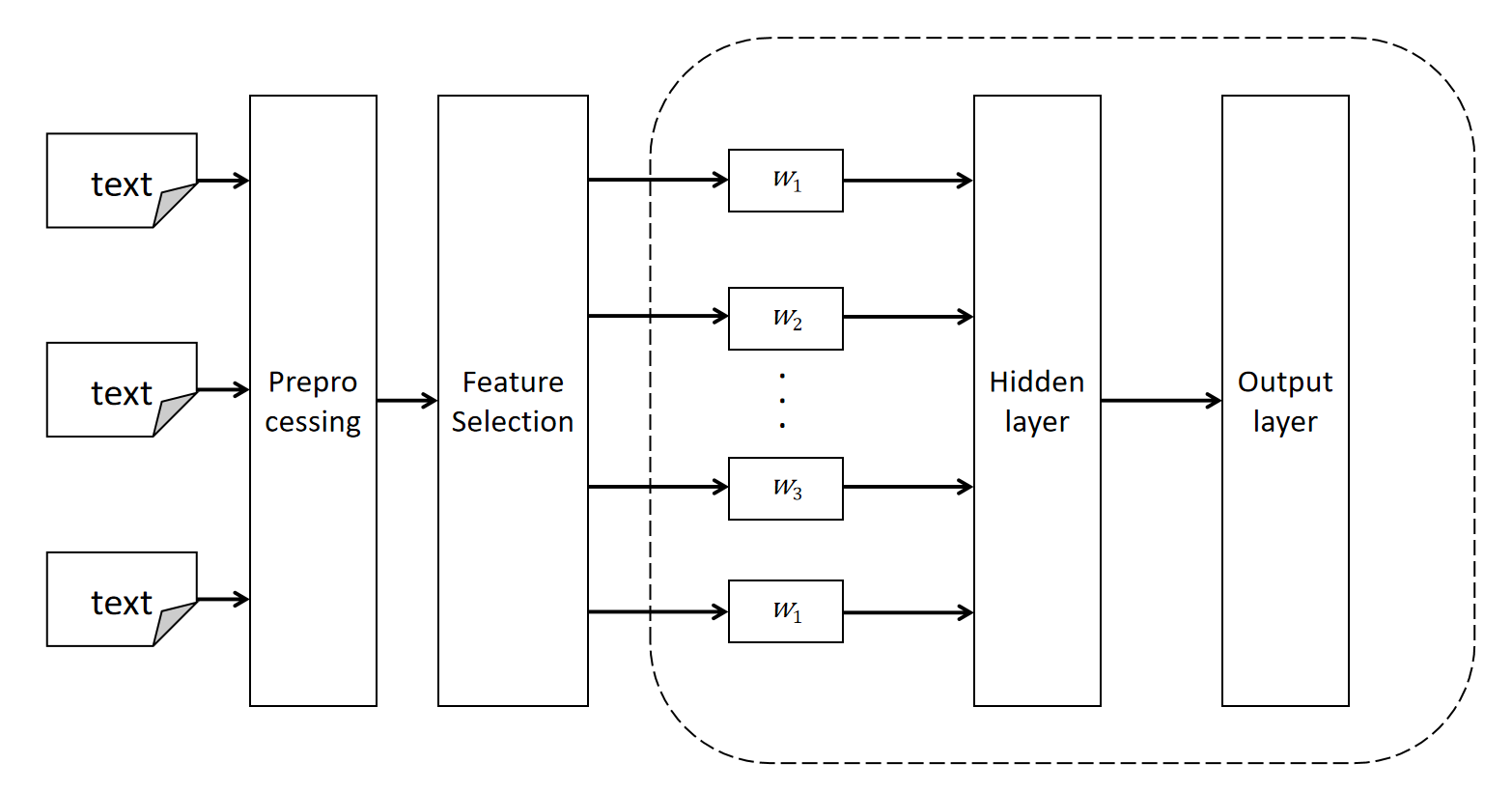}
	\end{center}
	\caption{Fasttext text sentiment analysis process. The input texts will be processed first,select features then.The features will be sent to hidden layer.And eventually output.}
\end{figure*}

The specific process of the FastTextt algorithm is as follows:
Input: preprocessed dataset.
Output: word vector $w_i$ and the probability $P_i$ of each comment in the dataset belonging to category i.

(1) Preprocess the target dataset, that is, remove emoticons, Chinese word segmentation, and stop words, and then process the preprocessed data according to the format of the model. Sentiment classification based on FastText is supervised, and labels are required to distinguish each comment.

(2) According to the research needs, choose fasttext.superviesd() for classification, or fasttext.skipgram() or fasttext.cbow() for generating word vectors.

(3) Train the processed data to obtain the corresponding word vector $w_i$.

(4) After processing the data set, according to the vector corresponding to the word and the vector corresponding to the n-gram word, obtain the probability $P_i$ of each comment belonging to category i.

\section{Model}
In this section, we will introduce the measure of the BERT-fasttext model. This model is based on BERT and fasttext, integrating parts of them to improve performance.

\subsection{\textbf{Keyword extraction algorithm}}
The extraction of syntactic structure phrase rules is based on the extraction of keywords in cyber violence language texts. The extraction effect of text keywords will directly affect the accuracy of establishing rules. Therefore, it is very necessary to find a suitable keyword extraction algorithm for cyber violence language texts.
There are three main directions for keyword extraction. One is a statistical method, including word frequency, TF-IDF, etc.; the second is a machine learning method, including support vector machine, conditional random field, etc.; the third is a semantic method, including part of speech, grammar, etc.
Based on the characteristics of the syntactic structure phrase form of cyber violence language, this paper combines statistical and semantic methods on the basis of $\chi^2$ statistical feature extraction, and proposes a new keyword extraction algorithm
$\chi^2$ - FPN. This algorithm is used as the calculation basis for keyword extraction. The specific parameters used are x1 2 statistics, word frequency, part of speech, word position.The function is below.
\begin{equation}
	\chi^{2}-F P N\left(C_{i}, D_{t}\right)=\chi^{2}\left(C_{i}, D_{t}\right) * \operatorname{fre}\left(C_{i}\right) * \operatorname{nom}\left(C_{i}\right) * \operatorname{pos}\left(C_{i}\right)
\end{equation}

Among them,$\chi^{2}\left(C_{i}, D_{t}\right)$is the chi-square statistic of word C and text category D,$\operatorname{fre}\left(C_{i}\right)$represents the frequency feature value of word $C_i$,$\operatorname{nom}\left(C_{i}\right) $represents the part-of-speech feature value of word $C_i$,$\operatorname{pos}\left(C_{i}\right)$represents the word position feature value of word $C_i$.
The word frequency feature value examines the proportion of a word in the category to which it belongs. This article classifies words based on part of speech, so the calculation formula for the word frequency feature value is as follows.
\begin{equation}
	\operatorname{fre}\left(C_{i}\right)=\frac{f(C_i)}{{\textstyle \sum_{n}f(C_i)}}
\end{equation}

Where: $f (Ci)$ represents the frequency of occurrence of word $C_i$, and $n$ represents the total number of words in the word category to which word $C_i$ belongs. The greater the number of times a word appears in the category to which it belongs, the greater its frequency feature value is considered.
\subsection{\textbf{Combination of rules and language models}}
The extracted rules may identify some non-violent language word combinations as violent language through the rule-based network violent language detection method, because the rules will treat the words in the rule dictionary indiscriminately, and some word combinations that meet the rules are not violent language. In order to solve this problem and reduce the false detection rate, this paper proposes a method that combines language models and rules, adds a constraint to the rules through the language model, and further optimizes the extracted rules.
\begin{equation}
	P(S)=P(W_1,W-2,...,W_t)=\prod_{i}^{t=1} P(W_i|Context)
\end{equation}
The word represented by the $n-gram$ model is only related to the n-1 words before it. For sentence $S$, its language model is expressed as follows.
\begin{equation}
	P(S) = P(W_1, W_2, \dots, W_t) = \prod_{i=1}^t P(W_i | W_{i-n+1}, W_{i-n+2}, \dots, W_{i-1})
\end{equation}
When n=1, the model is context-independent. The word only considers its own probability and relies solely on the word frequency statistics of the text, which does not have much practical application value. So when talking about n-gram, the value of n is generally n $\ge$ 2. The n-gram language model includes the information of the first n-1 words of the word, which has a strong constraint on the current word. Therefore, to date, the n-gram language model is the first choice in the practical application of many language models.

The n-gram model also has some limitations. Due to the problem of corpus, n-gram cannot train higher-order language models, that is, the value of n cannot be too large. At present, most research work or applications use Tri-gram or Four-gram, and Bi-gram is also used in some specific occasions. Another problem is that the n-gram model cannot establish the similarity between words.

\subsection{\textbf{Way of selecting features}}
Before representing the text as a vector, it is necessary to first select the words that can be used as text vector features. This is a very important step. Whether you can select representative and excellent feature words from the text for classification will largely determine the quality of the final classification effect.Sometimes we use the value below.

(1)Document Frequency,which Refers to the frequency of documents containing a certain feature item in the text set.

(2)Mutual Information,which is a criterion for measuring the correlation between two vectors.
\begin{equation}
	MI(t_i,c_j) = \log{\frac{p(t_i|c_j)}{p(t_i)}}=\log{\frac{p(t_i,c_j)}{p(t_i)\times{p(c_j)}}}
\end{equation}
\begin{equation}
	I(x,X) = \sum_{y} p(y |x) \log{\frac{p(y|x)}{p(y)}}
\end{equation}
In another way,you can use this function to count MI value.
\begin{equation}
	MI(t_i,c_j) = \log{\frac{{A}\times{D}}{(A+B)\times(A+C)}}
\end{equation}

(3)Information Gain,which is the amount of information provided by whether a feature item appears in a document to determine which category the document belongs to.
\begin{equation}
	IG(t_i) = \sum_{j = 1}^{k} P(c_j) \times \log P(c_j) + P(t_i) \times \sum_{j = 1}^{k} P(c_j|t_i) \times \log P(c_j|t_i) + P(\overline{t_i}) \times \sum_{j = 1}^{k} P(c_j|\overline{t_i}) \times \log P(c_j|\overline{t_i})
\end{equation}

(4)Chi-square,A method to measure the correlation between feature item $t$ and category $c$.
\begin{equation}
	\chi^{2}(t,c) = \frac{N \times (AD - CB)^{2}}{(A + C)(B + D)(A + B)(C + D)}
\end{equation}

\section{Experiments}
In this section, we first introduce our experimental settings, including datasets and evaluation metrics. Then, we provide comparative analysis with state-of-the-art methods on various benchmark datasets. Finally, we give a self-evaluation and future prospects of our proposed model.
\subsection{\textbf{xperiment setting}}
\textbf{Dataset.} 
In this paper, we will conduct the experiment on the following several datasets including  HateSpeechDataset.This dataset contains 440,906 data, with the features "Content", "Label", "Content\_int" and the type "object".The following is an example.

\begin{table}[H]
	\caption{ dataset of HateSpeechDataset}
	\begin{center}
		\footnotesize
		\begin{tabular}{m{5cm} c m{8cm}}
			\hline
			Content &  Label & Content\_int  \\
			\hline
			denial of normal the con be asked to comment on tragedies an emotional retard  &1 & [146715, 0, 1, 2, 3, 4, 5, 6, 7, 8, 9, 10, 11, 12, 13, 146714] \\ 
			just by being able to tweet this insufferable bullshit proves trump a nazi you vagina &1 &[146715, 14, 15, 16, 17, 7, 18, 19, 20, 21, 22, 23, 24, 25, 26, 27, 146714] \\ 
			king eric canton at manchester united eric canton is one of the best foot ball players of all time he scored total goals &0 &[146715, 629, 10835, 7517, ..., 3085, 197, 474, 948, 568, 177, 4558, 316, 719, 146714] \\ 
			\hline
		\end{tabular}
	\end{center}
	\label{tabdsa}
\end{table}

\subsection{\textbf{Evaluation metrics}}
In order to comprehensively and accurately evaluate the performance of our proposed brute force text recognition model combining BERT and fasttext, we selected the following four key evaluation indicators: $\textbf{Accuracy}$,$\textbf{Precision}$, $\textbf{Recall}$, and $\textbf{F1\_Score}$. These indicators can reflect the performance of the model from different angles and help us understand the strengths and weaknesses of the model more deeply.\cite{Ren2024MultimodalSA}

\begin{equation}
	Acc = \frac{TP + TN}{TP + TN + FP + FN}
\end{equation}

\begin{equation}
	Pre = \frac{TP}{TP + FP}
\end{equation}

\begin{equation}
	Recall = \frac{TP}{TP + FN}
\end{equation}

\begin{equation}
	F1\_score = \frac{2 \cdot Pre \cdot Recall}{Pre + Recall}
\end{equation}

\subsection{\textbf{Contrast experiment}}
In order to gain a deeper understanding of the performance advantages of the brute force text detection model combined with BERT and fasttext, and to verify its applicability in different scenarios, we designed a series of comparative experiments. These experiments aim to fully evaluate the performance of our proposed model by comparing different models and different feature combinations. The experimental results are shown in Table 2.
\begin{table}[H]
	\caption{Evaluation results of different models and different feature combinations}
	\begin{center}
		\footnotesize
		\resizebox{\textwidth}{!}{
			\begin{tabular}{m{4.5cm}m{2.5cm}m{2.5cm}m{2.5cm}m{2.5cm}}
				\hline
				Model &  Acc (/\%) & Pre (/\%) & Recall (/\%) & F1\_score (/\%) \\
				\hline
				fasttext & 86.8 & 86.4 & 86.9 & 86.0 \\ 
				BERT & 86.9 & 87.3 & 86.8 & 87.0 \\
				\textbf{BERT\_fasttext} & \textbf{87.6} & \textbf{87.4} & \textbf{87.6} & \textbf{86.6} \\
				word2vec & 82.0 & 75.0 & 82.0 & 74.0 \\
				ResNet & 86.9 & 66.9 & 61.3 & 64.0 \\
				\hline
			\end{tabular}
		}
	\end{center}
	\label{tabdsa}
\end{table}

\subsection{\textbf{Results Analysis}}
From the above experimental results, it can be seen that the best performance on the dataset is achieved by using the BERT\_fastetext Model.\cite{Feng2024AudiosDL}

\textbf{Acc}:BERT\_fasttext has an accuracy of 87.6\%, which means that the model has a high percentage of correct predictions in all test samples. A high accuracy indicates that the model is doing well overall, and BERT\_fasttext performs very well compared to other models.

\textbf{Precision}:The precision is 87.4\%, which means that among all the samples predicted as positive by the model, 87.4\% are true positive samples. A high precision indicates that the model has a low false positive rate when predicting the positive class.

\textbf{Recall}:The recall rate is 87.6\%, which means that among all the samples that are actually positive, the model can correctly predict 87.6\% of the positive samples. A higher recall rate indicates that the model can better capture positive samples and reduce false negatives.

\textbf{F1\_socre}:The F1-score is 86.6\%, which is the harmonic mean of precision and recall. A high F1-score indicates that the model has found a good balance between precision and recall.

Even though our model performs well on this dataset, we can still improve it. We can target the social network Sina Weibo, crawl the comment text data of cyber violence incidents, and obtain a dataset for research through data denoising and text preprocessing. We can use semi-supervised learning methods, use a small amount of manual intervention and multiple generations to select features, and finally construct a high-quality cyber violence language corpus to fill the gap in the Chinese cyber violence language corpus. Then, we can train the Chinese dataset based on this model.

\section{Conclusion}
This study proposed a text detection method based on the fusion of BERT and fasttext models, explored the detection ability of violent text, and aimed to combine the text understanding ability of BERT with the efficiency of fasttext to improve the accuracy and robustness of text detection. Experimental results show that this model successfully achieved the best text detection effect on standard datasets. Specifically, the BERT model provides accurate text representation for sentiment analysis through its powerful language understanding ability, while using FastText for fast text classification has advantages in computational efficiency and reasoning speed. The contribution of this study is to combine BERT with fasttext models for violent text detection tasks, which provides new ideas for research and application in the field of text detection, and injects new vitality into the field of violent text. Future work can further explore more complex deep learning technologies and their application effects in practical scenarios.



\quad


\bibliography{mybibfile}

\end{document}